\documentclass[runningheads]{llncs}
\usepackage{multirow,multicol,booktabs,graphicx,CJKutf8,color,makecell,tabularx} 
\usepackage{algorithm,algorithmic,amsmath}
\usepackage{bbding}
\usepackage{subfigure}  

\begin{document}
%
\title{CPTuning: Contrastive Prompt Tuning for Generative Relation Extraction}

\author{Jiaxin Duan\orcidID{0000-0001-6137-6632} \and
Fengyu Lu \and
Junfei Liu}
\authorrunning{Duan et al.}
\institute{
Peking University, Beijing, China
\\
\email{\{duanjx,fengyul\}@stu.pku.edu.cn}, 
\email{liujunfei@pku.edu.cn}
}


%
\maketitle
\begin{abstract}
Generative relation extraction (RE) commonly involves first reformulating RE as a linguistic modeling problem easily tackled with pre-trained language models (PLM) and then fine-tuning a PLM with supervised cross-entropy loss.
Although having achieved promising performance, existing approaches assume only one deterministic relation between each pair of entities without considering real scenarios where multiple relations may be valid, i.e., entity pair overlap, causing their limited applications. 
To address this problem, we introduce a novel contrastive prompt tuning method for RE, CPTuning, which learns to associate a candidate relation between two in-context entities with a probability mass above or below a threshold, corresponding to whether the relation exists. 
Beyond learning schema, CPTuning also organizes RE as a verbalized relation generation task and uses Trie-constrained decoding to ensure a model generates valid relations. 
It adaptively picks out the generated candidate relations with a high estimated likelihood in inference, thereby achieving multi-relation extraction. 
We conduct extensive experiments on four widely used datasets to validate our method. Results show that T5-large fine-tuned with CPTuning significantly outperforms previous methods, regardless of single or multiple relations extraction.

\keywords{Relation extraction \and Prompt tuning \and Contrastive learning.}
\end{abstract}

\section{Introduction}
\label{sec:typesetting-summary}
\begin{figure}[!t]
\centering
\includegraphics[width=0.55\textwidth]{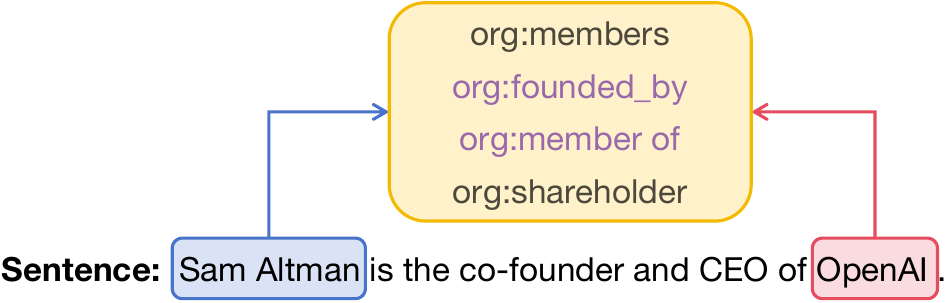}
\caption{The standard paradigm of relations extraction.}
\label{fig:1}
\end{figure}

Relation extraction (RE) is one of the fundamental tasks in natural language processing (NLP), aiming to extract relational facts from unstructured text into structured triplets~\cite{survey,survey1}. Figure~\ref{fig:1} illustrates the standard RE paradigm: given an instance that includes a sentence containing two identified entities, a model is required to categorize the relationship between them from a predefined set of relations. 
Human knowledge supported by RE significantly boosts the development of downstream knowledge-sensitive applications, such as dialogue systems~\cite{DBLP:conf/emnlp/ZhaoWXTZY20}, question answering~\cite{QA-GNN}, and knowledge graph completion~\cite{TuckER}. 

Recently advanced methods for relation extraction can be roughly recognized into two styles: classificational style and generative style. 
Both styles were recently developed based on pre-trained language models (PLMs)~\cite{bart,t5} and involved using handicraft templates to transform RE into a fundamental linguistic problem to evoke pre-training potential, namely \textit{prompt tuning}~\cite{prompt-survey}. For example, classificational FPC~\cite{FPC} reformats RE as relation words prediction to align with the masked language modeling (MLM) pre-training objective~\cite{BERT}, while generative GenPT~\cite{GenPT} converts RE to text-infilling to coordinate with sequence-to-sequence (Seq2Seq) MLM~\cite{bart}. 
Compared with traditional methods that predict numerical relation labels depending on contextual features, prompt tuning methods preserve the semantics of the relation described in text form and, therefore, have achieved unprecedented performance. Despite that, almost all existing studies assume only one deterministic relation over each entity pair beforehand without considering the multiple correct relations existence.

The case in RE that more than one relation links a pair of entities is referred to as entity pair overlap (EPO)~\cite{survey,survey2}. Taking the sentence “Sam Altman is the co-founder and CEO of OpenAI.” as an example, two relations are established between the entities “Sam Altman” and “OpenAI,” i.e., (Sam Altman, org:founded\_by, OpenAI) and (Sam Altman, org:member\_of, OpenAI). prompt tuning methods tend to fabricate a template like “In this sentence, the relation between [Sam Altman] and [OpenAI] is [MASK] ... [MASK] sentence:”, where [MASK] tokens sit the slots of relation words to be predicted by a PLM. 
On the one hand, maximum likelihood estimation (MLE) is the most used objective to tune such a model, which assigns probability mass 1 to the gold relation while 0 to all other potential ones. 
This deterministic assumption intrinsically contradicts EPO and heavily restricts the use of ranking-based prediction (see Section 4).
Intuitively, placing more slots into the template can extend MLE learning to the EPO scenario. However, the number of slots is hard to determine because of an agnostic number of valid relations described in varying words. As a result, the shortcoming of prompt tuning-based RE methods - unworkability for EPO significantly limits their applications.

In this paper, we propose CPTuning, a novel Contrastive Prompt Tuning paradigm for RE that facilitates generative PLMs to overcome EPO and acquire better performance. 
Specifically, CPTuning reforms RE into Seq2Seq text-infilling. It uses a textual template to present RE samples in corrupted neural language, where the words about entities and their relations are masked. 
With such a synthetic sample as input, CPTuning learns a Seq2Seq PLM to generate a verbalized candidate relation by the probability higher or lower than a predefined threshold, depending on whether the relation truly exists. 
Notably, a label smoothing trick~\cite{label-smoothing} is used during learning, which softens the one-hot relation word distributions to ensure the model estimates any candidate relation by a non-zero probability.
Besides, a Trie~\cite{SURE} is built conditioned on the given relation set to constrain the model estimation concerns only valid relation words. 
At inference time, CPTuning further hires a prefix-given beam search~\cite{beam-search} strategy to sample multiple candidate relations from the model outputs, and only those with an estimated score above a certain borderline are deemed extraction results.
In this way, EPO is effectively coordinated through adaptive diverse relations generation.

Our main contributions are as follows: 
\begin{itemize}
\item We propose to reform RE into Seq2Seq text-infilling and adopt a beam search strategy with Trie-constraints to generate diverse relations adaptively.
\item We propose CPTuning, which learns a generative RE model to assign relatively higher or lower probability mass to a candidate relation depending on whether it is established.
\item Extensive experiments on four public benchmarks show that CPTuning enables a PLM to deal with EPO effectively and achieve state-of-the-art results on single and multiple relations extraction. In-depth analyses further attribute the CPTuning advanced performance to the capture of semantic information in verbalized relation labels.
\end{itemize}

\section{Related Work}
\subsection{Relation Extraction}
Having seen the great success of pre-training technology, researchers adapted widely PLMs for RE in recent years. The earlier works~\cite{RECENT,R-BERT} treated RE as a traditional discriminate task and focused on fine-tuning PLMs to capture rich semantic features of entity pairs and their context that are then fed into an additional classifier for multi-class classification. 
Since prompt tuning (see in next subsection) has shown remarkable potential on various NLP tasks, the latest methods began to use handcraft templates to transform RE into easier-handle problems. We can roughly divide them into two groups, namely classificational-style and generative-style. During pre-processing, both styles verbalize numerical relation labels as brief textual descriptions. 
Classificational methods~\cite{FPC} commonly use cloze templates to organize the RE instance (including highlighted entity pair) to align with the MLM task format. Generative methods~\cite{GenPT,GenIE} may also adopt a cloze template to reformat the instance but require a PLM to generate verbalized relation labels in an autoregressive manner. Beyond reforming RE as foundational linguistic tasks used for pre-training, some works convert RE into summarization~\cite{SURE}, reading comprehension~\cite{DBLP:conf/conll/LevySCZ17}, and machine translation~\cite{DBLP:conf/iclr/PaoliniAKMAASXS21} with elaborate templates to conveniently induce the relevant knowledge contained in the upstream models that benefit RE. 

\subsection{Prompt Tuning}
Prompt-based fine-tuning is an increasingly popular transfer learning strategy, which unlocks the PLMs' potential by using learnable or handcraft templates that align downstream tasks with the pre-training objectives~\cite{prompt-survey}. 
According to the template format, prompt tuning methods are classified as cloze prompt tuning and prefix prompt tuning. Cloze prompt~\cite{FPC,DBLP:conf/acl/CuiWLYZ21} adopts the templates with two slots for individually filling the input and target texts\footnote{In the text classification task, a verbalizer is additionally required to map numerical labels as language words.}. By contrast, prefix prompt~\cite{prefix-tuning,prompt-tuning,p-tuning} concatenates the input and target texts as a single and further prepends a few words ahead of it. The combination of PLMs plus cloze prompt tuning is widely used in RE and has shown remarkable performance. However, limited by MLE learning and inflexible handcrafted templates, existing methods struggled with the EPO challenge. In this paper, we follow the current frequently used templates and introduce a novel contrastive learning framework, which makes our CPTuning cope with EPO effectively.

\begin{figure}[!tp]
\centering
\subfigure[Handcrafted templates used for formalizing RE samples.]{
\includegraphics[width=1.0\linewidth]{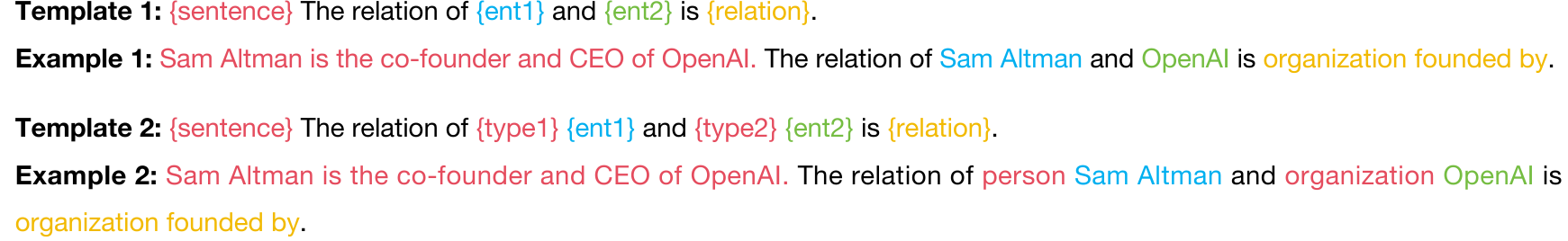}
\label{fig:2a}
}
\subfigure[The syntax structure of source and target texts in our transformed RE task.]{
\includegraphics[width=1.0\linewidth]{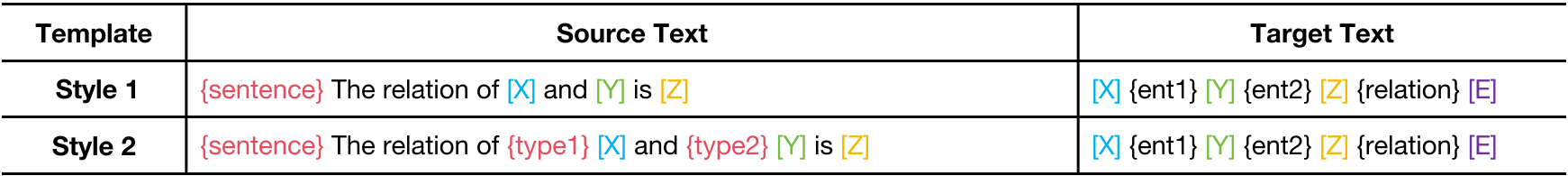}
\label{fig:2b}
}
\caption{
The syntax structure of (a) our handcrafted templates and (b) the synthetic source and target texts in the transformed RE task.
\{sentence\}: instance text slot, \{entX\}: entity slots, \{relation\}: relation words slot, and \{typeX\}: entity type slots.
Sentinel tokens [X], [Y], and [Z] corrupt an original formalized instance to build a source text.
[E] indicates the end of a target text.
}
\label{fig:2}
\end{figure}

\section{Method}
In this section, we present the details of CPTuning, a contrastive prompt tuning paradigm that adapts Seq2Seq PLMs for RE and overcomes the EPO challenge. 
We transform RE into Seq2Seq text-infilling in Section~\ref{sec:3.1}, describe the generation and scoring of candidate relations in Section~\ref{sec:3.2}, and illustrate the contrastive-based learning procedure in Section~\ref{sec:3.3}.

\subsection{Task Definition}
\label{sec:3.1}
Let $\mathcal{D}=\{\mathcal{X},\mathcal{Y}\}$ present a RE dataset, including the instance set $\mathcal{X}$ and the relation set $\mathcal{Y}$. 
In general paradigm, given an instance text $x_{i} \in \mathcal{X}$ with two highlighted entities $e^{i}_1$ and $e^{i}_2$, the objective of RE is to predict these two entities' relation $y_i \in \mathcal{Y}$. 
Existing works always view RE as a multi-class classification (without considering EPO) or multi-label classification (consider EPO) task, modeling the label distribution $\mathrm{P}(y|x_{i},e^{i}_1,e^{i}_2), y \in \mathcal{Y}$. 

\begin{figure}[!tp]
\centering
\includegraphics[width=1.0\textwidth]{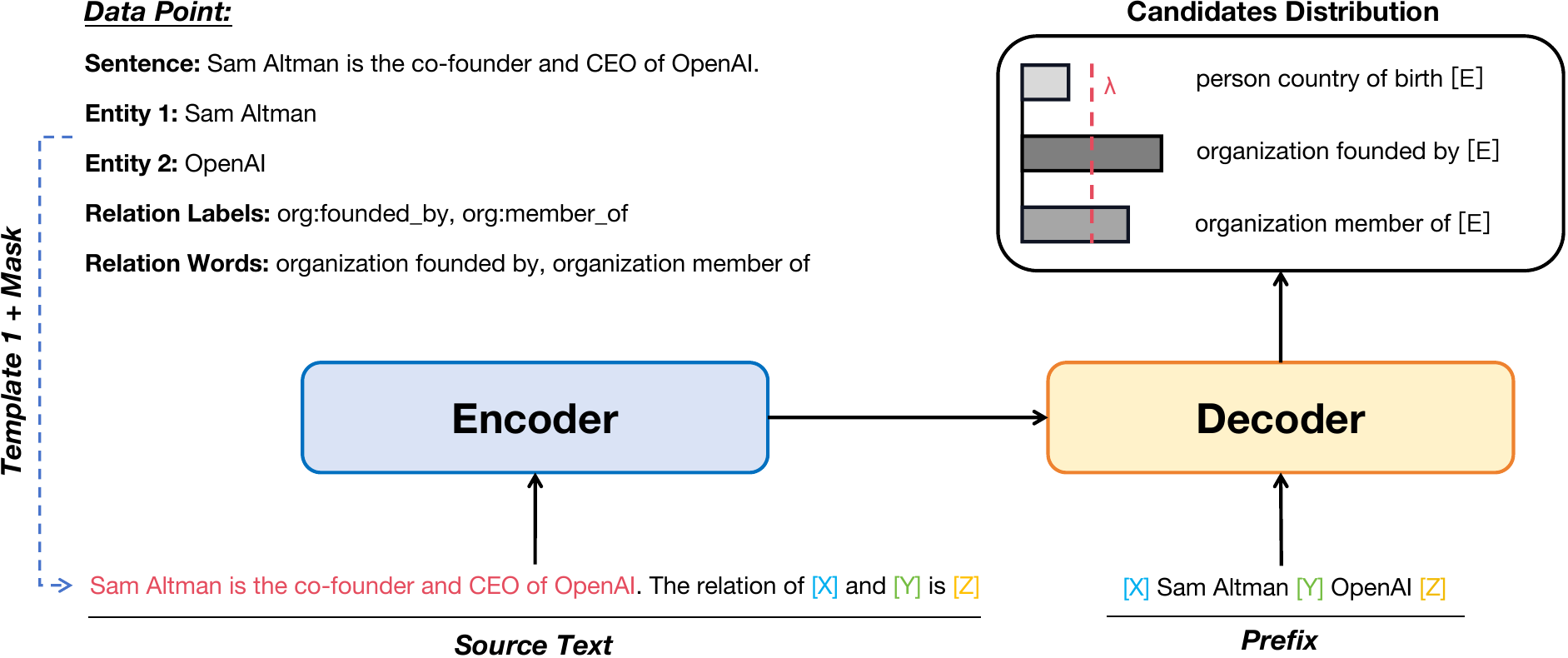}
\caption{The illustration of the Seq2Seq text-infilling task transformed from RE.}
\label{fig:3}
\end{figure}

Intuitively, any classifier like multi-layer perceptron (MLP) plus a softmax operator can be used to solve the above classification problems. However, to preserve semantics information, which has been confirmed beneficial in RE, we consider encoder-decoder PLMs and reformulate RE as a text-infilling task. 
For this reason, we first implement a verbalizer $v(\cdot)$ following \cite{GenPT} to rewrite relation labels with short phrases. For example, the label “org:top$\_$members/employees” is presented in words “organization top members or employees,” and “no$\_$relation” is written as “no relation.” 
Then, we use cloze templates to present instance texts and the corresponding verbalized relation labels (called 'relation words' in the following) into natural language. As shown in Figure~\ref{fig:2a}, we introduce two types of templates with slight differences in syntactic structure. Specifically, the first one contains sentence slots and entity slots for placing the instance text and entities, respectively, and the second provides additional slots for asserting the type of the succeeding entity. 
We denote the template as a function $h(\cdot,\cdot)$ for illustration convenience, and Section~\ref{sec:4.3} will show that the different prompting content in the two templates results in discrepant learning efficacy.

After verbalization and templating, a formalized RE sample $s=h(x,v(y))$ meets the preference of PLMs in processing language text. Still, this format is model-agnostic, i.e., not intentionally designed to fit the advantage of a specific model (or a class of models). Inspired by previous work~\cite{SURE} that transformed RE into abstractive summarization, we transform RE into Seq2Seq text-infilling to align with the pre-training task of the advanced T5~\cite{t5}. 
Specifically, taking $s$ as an original text, we introduce sentinel tokens to mask its slots of entity, entity type, and relation at the source end and induce the recovery of these slots at the target end.
The syntactic structure of the resulting source and target texts are listed in Figure~\ref{fig:2b}, depending on template styles. 
We also demonstrate in Figure~\ref{fig:3} the transformed RE task, which aims to model the probability of a synthetic target text $t$ conditioned on a corrupted sample, i.e., $\mathrm{P}(t|\mathrm{M}(s))$, where $\mathrm{M}(\cdot)$ denotes the text corruption with sentinel tokens. 
 

\begin{figure}[!t]
\centering
\includegraphics[width=1.0\textwidth]{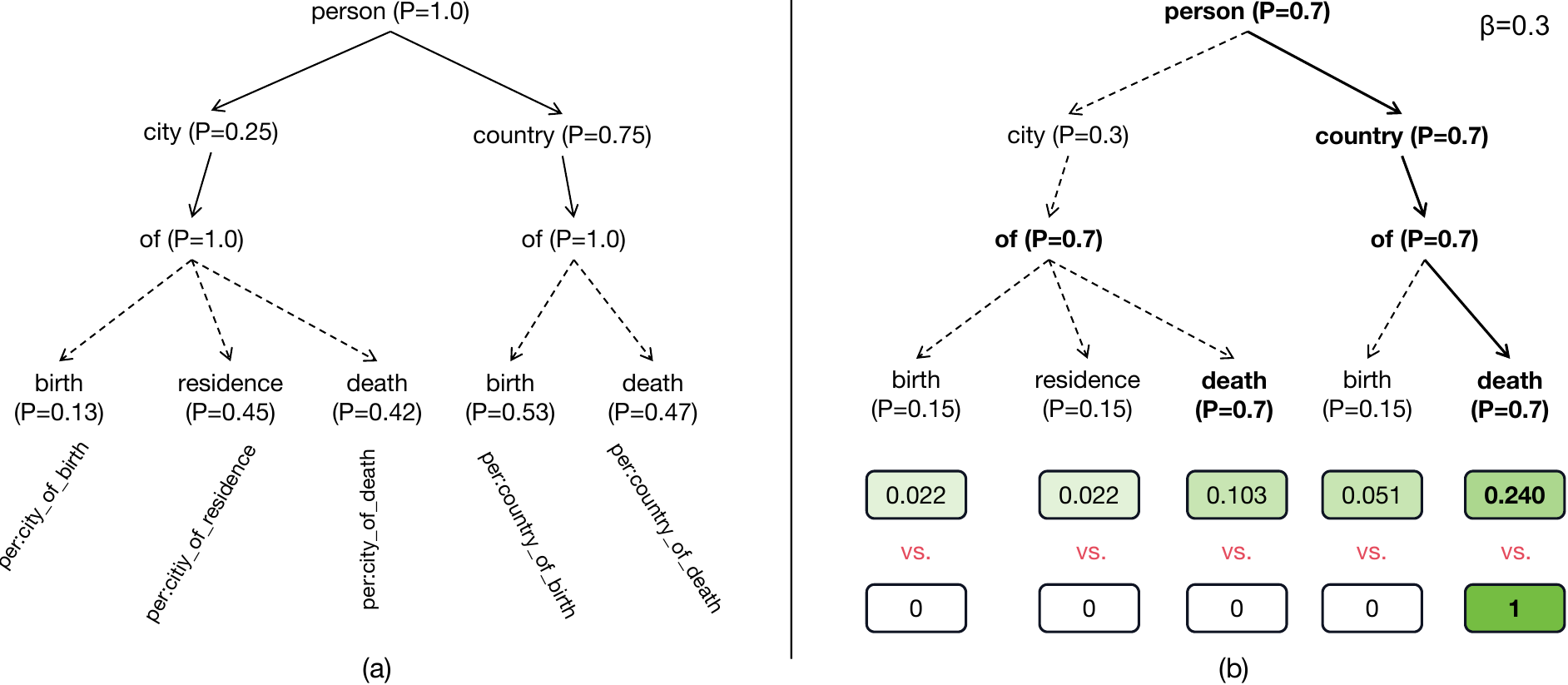}
\caption{
The illustration of Trie, PGC decoding (a), and LBLS (b). 
The five relations predefined beforehand are "per:city\_of\_birth", "per:citiy\_of\_residence", "per:city\_of\_death", "per:country\_of\_birth", and "per:country\_of\_death", respectively.
We contrast the probability mass of sampling a candidate relation with and without LBLS in the (b) bottom.
}
\label{fig:4}
\end{figure}

\subsection{Candidate Relations Generation and Scoring}
\label{sec:3.2}
It is worth noting that since we assume more than one relation is established between two entities in a given RE instance, any candidate relation estimated by a relatively large probability should be checked for existence.
Therefore, how to access the estimated likelihood of a candidate relation in our paradigm is considered. 
On the one hand, although we mask three types of slots in the source text, only the recovered relation slot in the target text is necessary. Moreover, each token of the generated target text is conditionally sampled from the PLM vocabulary, causing a risk of producing invalid relation words. We draw ideas of \cite{GenIE,SURE,GenPT} and perform prefix-given constrained decoding (PGC) at inference time to overcome these two challenges. 
To this end, we divide a target text into dual parts: prefix $z$, including words before the relation slot, and relation $r$, including words after the sentinel [Z], and the probability $\mathrm{P}(t|\mathrm{M}(s))$ can in turn be unfolded:
\begin{equation}
\label{eq:1}
\mathrm{P}(t|\mathrm{M}(s))=\mathrm{P}(r|\mathrm{M}(s),z) \cdot \mathrm{P}(z|\mathrm{M}(s))
\end{equation}

Reviewing in Figure~\ref{fig:2b} that all candidate relations regarding a certain instance share an identical prefix, we thus omit the second term on the right side of Eq.\ref{eq:1} and solely model the relation part with pre-trained parameters $\theta$:
\begin{equation}
\label{eq:2}
\mathrm{P}(r|\mathrm{M}(s),z;\theta)=\prod_{i=1}^{|r|} \mathrm{P}(r_{(i)}|\mathrm{M}(s),z,r_{(<i)};\theta)
\end{equation}
where $r=\{r_{(1)},r_{(2)},\cdots,r_{(|r|)}\}$ and $r_{(i)}$ is a word.

\subsubsection{Prefix-given Constrained Decoding}
Our PGC decoding refers to sampling the $K$ most potential candidate relations $r_1,r_2,\cdots,r_K$ from PLM outputs conditioned on a known prefix. It is very similar to the standard beam search (BS)~\cite{beam-search}, which retains at most $K$ branches on each decoding step, except that the first token sampled with PGC is conditioned on prefix text rather than a BOS (Beginning of Sentence) token. Besides, the searching space of PGC is constrained by a Trie~\cite{SURE}, where the possible next token in a branch can only be one of the children of the leaf node, depending on the relation set. For example, according to the five relations given in Figure~\ref{fig:4}a, the next token of the incomplete phrase “person country of” is restricted to determine between “birth” and “death.” 
Compared with searching on the model vocabulary, Trie-based searching offers greater efficiency and makes sure the sampled relation words are definitely valid. 

\begin{algorithm}[!t]
\renewcommand{\algorithmicrequire}{\textbf{Input:}}
\renewcommand{\algorithmicensure}{\textbf{Output:}}
\small
\caption{Prefix-given Constrained Inference Algorithm}
\begin{algorithmic}[1]
\REQUIRE
A prefix $z$, an input instance $\mathrm{M}(s)$, and a predefined Trie with $N$ layers
\ENSURE
A candidate relation set $\mathcal{R}$
\STATE Initialize $K$ verbalized candidate relations $r_1=z,r_2=z,\cdots,r_K=z$ \\
\FOR{$i \gets 1,N$}
    \STATE $\mathcal{R}=\{ \}$
    \FOR{$j \gets 1,K$}
        \STATE Search in the Trie to get the leaves of the branch $r_j$, denoted as $\mathcal{L}$
        \FOR{$k \gets 1,|\mathcal{L}|$}
            \STATE $\mathcal{R}=\mathcal{R} \cup \{ r_j + \mathcal{L}_k\}$
        \ENDFOR
    \STATE Reset $r_1,r_2,\cdots,r_K$ as the top-K elements in set $\mathcal{R}$ that achieve the highest probability mass estimated by Eq.~\ref{eq:2}
    \ENDFOR
\ENDFOR
\STATE $\mathcal{R}=\{ \}$
\FOR{$i \gets 1,K$}
    \IF{$f(r_i)>\lambda$}
        \STATE $\mathcal{R}=\mathcal{R} \cup \{ r_i \}$
    \ENDIF
\ENDFOR
\RETURN $\mathcal{R}$
\end{algorithmic}
\label{algorithm:1}
\end{algorithm}

\subsubsection{Candidates Relation Scoring and Selection}
Once multiple candidate relations are sampled out by PGC decoding, we score each and retain only the most likely established ones as the final results. In this step, the length-normalized log-likelihood is considered:
\begin{equation}
\label{eq:3}
f(r_i)=\frac{-\sum_{j=1}^{|r_i|} \log \mathrm{P}(r_{(j)}^{i}|\mathrm{M}(s),z,r_{(<j)}^{i};\theta)}{|r_i|^{\alpha}}
\end{equation}
where $\alpha$ is a length penalty suggested in \cite{brio}. 
We also introduce a borderline $\lambda$; if $f(r_i)>\lambda$, the relation $r_i$ is regarded established otherwise, inexistent.
The overall decoding and scoring process is described in Algorithm 1 for clear understanding.

\subsection{Contrastive-based Learning}
\label{sec:3.3}
According to Eq.~\ref{eq:2}, without considering EPO, our goal of fine-tuning the PLM is to maximize the conditional probability of gold relation words (plus a sentinel token [E]). This is commonly known as maximum likelihood estimation (MLE), which trains a model to minimize the following cross-entropy loss:
\begin{equation}
\label{eq:4}
\mathcal{L}_{ce}(\theta)=-\sum_{i=1}^{|r|} \log \mathrm{P}(r_{(i)}|\mathrm{M}(s),z,r_{(<i)};\theta)
\end{equation}
When it comes to the EPO scenario, this approach causes twofold problems. In training, given a sample with more than one gold relation, it is undetermined which one to be maximized. 
A step further, the fine-tuned model ideally predicts one of the gold relations by the probability of 1 while others by 0, significantly contradicting our scoring strategy.
To make the model aware of the relative correctness of a candidate relation, we apply label smoothing~\cite{label-smoothing} on Eq.~\ref{eq:4}:
\begin{equation}
\label{eq:5}
\mathcal{L}_{lbls}(\theta)= -\sum_{i=1}^{|r|} \mathrm{P}^{*}(r_{(i)}^{*}|\mathrm{M}(s),z,r_{(<i)})\log \mathrm{P}(r_{(i)^{*}}|\mathrm{M}(s),z,r_{(<i)};\theta)
\end{equation}
which aims to fit the soft-label distribution:
\begin{equation}
\label{eq:6}
\mathrm{P}^{*}(r_{(i)}^{*}|\mathrm{M}(s),z,r_{(<i)}) = \left\{\begin{matrix}
 1-\beta,& r_{(i)}^{*}=r_{(i)} \\
\frac{\beta}{|\mathcal{T}_i|-1},& r_{(i)}^{*}\ne r_{(i)} \text{ and } r_{(i)}^{*}\in \mathcal{T}_i\\
0, & \text{otherwise}
\end{matrix}\right.
\end{equation}
where $r_{(i)}^{*}$ denotes a token the model predicted at the $i$-th step, and $\mathcal{T}_i$ collects the candidate words on the $i$-th layer of the Trie mentioned in Section~\ref{sec:3.2}. We call this technique \textit{layer-based label smoothing} (LBLS) and graphically illustrate it in Figure 4b, where the hyperparameter $\beta$ is set to 0.3. In this case, the Trie has four layers, and $N_1$ to $N_4$ are individually 1, 2, 1, and 4. It can be seen that LBLS trains a model to generate the gold relation words “person country of death” with a higher probability while generating others with a lower rather than directly overlooking them.

Additionally, to ensure the effectiveness of our scoring mechanism, we further propose a contrastive loss:
\begin{equation}
\label{eq:7}
\mathcal{L}_{ctl}(\theta)=\sum_{r_i \in \mathcal{R}_{s}^{+} }^{} \max (\zeta-f(r_i),0)+\sum_{r_j \in \mathcal{R}_{s}^{-} }^{} \max (f(r_j)-\zeta,0)
\end{equation}
where $\mathcal{R}_s$ denotes a candidate relation set w.r.t the sample $s$, containing gold relations $\mathcal{R}_s^{+}$, as well as fake relations $\mathcal{R}_s^{-}$ randomly sampled from $v(\mathcal{Y})$. 
In this way, the relative quality of probability masses the model assigns to the gold and negative relations are explicitly controlled by a threshold $\zeta$.
Finally, we follow \cite{brio} and combine the two types of losses with a balance factor $\mu$:
\begin{equation}
\label{eq:8}
\mathcal{L}(\theta)=\mathcal{L}_{ctl}(\theta) + \mu \mathcal{L}_{lbls}(\theta)
\end{equation}

\section{Experiments }
\subsection{Datasets}
\label{sec:4.1}
We use four well-known RE datasets to conduct experiments, including:

\textbf{TACRED}~\cite{DBLP:conf/emnlp/ZhangZCAM17} is one of the most popular RE datasets categorizing entities into subject and object classes and providing the spans and types of entities. The special case "no\_relation" is also considered.

\textbf{TACREV}~\cite{TACRED} is developed on the TACRED basis. It widely re-annotates the incorrect samples in the development and test sets while retaining the original training set.

\textbf{Re-TACRED}~\cite{Re-TACRED} is another revised version of TACRED that relabels the full dataset and rectifies a few relation labels with unclear meanings. 

\textbf{NYT}~\cite{NYT} is a well-known relation extraction dataset often used to evaluate a model in handling SEO (Single Entity Overlap) and EPO issues. The number of EPO entity pairs in the NYT training and test sets are 9782 and 987, respectively.

The detailed dataset statistics are presented in Table~\ref{tab:1}.

\begin{table}[!htbp]
\centering
\scriptsize
\caption{
The statistics of datasets.
\# counts the samples, relations, and overlapped entities in the training/development/test set.
}
\label{tab:1}
\begin{tabular}{c|p{40pt}<{\centering}|p{40pt}<{\centering}|p{40pt}<{\centering}|p{50pt}<{\centering}|p{40pt}<{\centering}}
\toprule
Dataset & \#Train & \#Dev & \#Test & \#Relations & EPO \\
\midrule
TACRED & 68,124 & 22,631 & 15,509 & 42 & No \\
TACREV & 68,124 & 22,631 & 15,509 & 42 & No \\
Re-TACRED & 58,465 & 19,584 & 13,418 & 40 & No \\
NYT & 56,195 & 5,000 & 5,000 & 24 & Yes \\
\bottomrule
\end{tabular}
\vspace{-10pt}
\end{table}

\subsection{Baseline Methods}
\label{sec:4.2}
We select the following representative methods across classificational and generative styles as baselines for comparison.
\textbf{SFT}, i.e., RoBERTa~\cite{roberta} or T5~\cite{t5} supervise fine-tuned with cross-entropy loss. For T5, the target text is our verbalized relation.
\textbf{GDPNet}~\cite{DBLP:conf/aaai/XueSZC21} searches indicative words to enhance relation representations.
\textbf{SpanBERT}~\cite{SpanBERT} learns to model spans to capture better structure information.
\textbf{MTB}~\cite{DBLP:conf/acl/SoaresFLK19} learns relation representations directly from the entity-linked text.
\textbf{KnowBERT}~\cite{DBLP:conf/emnlp/PetersNLSJSS19}, retrieving entity embeddings to update word representations.
\textbf{K-Adapter}~\cite{K-Adapter} injects multiple kinds of knowledge into a model with adapters.
\textbf{LUKE}~\cite{LUKE} treats words and entities as tokens to directly model relations.
\textbf{CR}~\cite{DBLP:conf/emnlp/ZhouC21} encourages several identical models to predict a similar likelihood.
\textbf{RECENT}~\cite{RECENT} restricts candidate relations using the constraint of entity types.
\textbf{TANL}~\cite{DBLP:conf/iclr/PaoliniAKMAASXS21} frames RE as a translation task based on augmented natural language.
\textbf{NLI}~\cite{sainz2021label} reformulates RE as a natural language inference task.
\textbf{SURE}~\cite{SURE} transforms RE into a text summarization task.
\textbf{PTR}~\cite{DBLP:journals/aiopen/HanZDLS22} decomposes RE into three subtasks and combines their results as the system output.
\textbf{KnowPrompt}~\cite{KnowPrompt} automatically builds templates and label words with knowledge.
\textbf{GenPT}~\cite{GenPT} converts RE to a text-infilling task using a template and label words.
\textbf{FPC}~\cite{FPC} converts RE into MLM and fine-tunes a model with curriculum-guided prompting.

\subsection{Implementation Details}
\label{sec:4.3}
We implement the standard version of CPTuning with T5$_{large}$\footnote{\url{https://huggingface.co/t5-large}} as the base model. 
During fine-tuning, we train the model for at most 10 epochs, and the batch size is set to 32.
We adopt an AdamW~\cite{loshchilov2017decoupled} optimizer with a linear learning schedule. 
The learning rate is initially set to 2.5e-5. It warms up during the first 10\% training steps and decays to 0 gradually in the subsequent steps.
All our experiments are conducted on one NVIDIA Tesla A100 GPU, and the codes are developed with Pytorch\footnote{\url{https://pytorch.org}} and Transformer\footnote{\url{https://huggingface.co/docs/transformers}} libraries. 
As for hyperparameter settings, we set $\alpha=0.6$ in Eq.~\ref{eq:3} and $\beta=0.2$ in Eq.~\ref{eq:6}. 
The borderline $\lambda$ is set to 1.0, the threshold $\zeta$ is set to 1.2, and the balance factor $\mu$ is set to 0.1.
Whenever the PGC beam search decoding is performed, the beam size $K$ is 16. We also sample 16 gold and fake relations during contrastive training represented by Eq.~\ref{eq:7}.
Following previous studies, we use micro F1 scores (\%) as the metric for model evaluation.

\begin{table}[!t]
\centering
\scriptsize
\caption{
Micro F1 scores on the test sets of the used datasets. 
Results marked by $\dag$ are from our reproduction, while others are reported in the original papers. 
The "Prompting" column illustrates whether a model is learned with prompt tuning. 
The "Base Model" column shows the adopted PLMs of corresponding methods.
In the "Extra Data" column, \textit{w/o} indicates the model only uses the data from the datasets, and \textit{w/} indicates the model is enhanced by incorporating additional data or knowledge bases.
\textit{-s1}: the first-style template without entity type slots. 
\textit{-s2}: the second-style template without entity type slots. 
The best results are in \textbf{bold}. 
}
\label{tab:2}
\begin{tabular}{c|c|p{45pt}<{\centering}|p{45pt}<{\centering}|p{40pt}<{\centering}|p{40pt}<{\centering}|p{50pt}<{\centering}}
\toprule
Methods & Base Model & Prompting & Extra Data & TACRED & TACREV & NYT \\
\midrule
\multicolumn{7}{c}{Classificational Methods} \\
\midrule
SFT & RoBERTa$_{large}$ & No & \textit{w/o} & 68.7$^{\dag}$ & 76.0$^{\dag}$ & 88.23$^{\dag}$ \\
GDPNet & SpanBERT$_{large}$ &  No & \textit{w/o} &  70.5 &  80.2 &  - \\
SpanBERT & BERT$_{large}$ & No & \textit{w/o} & 70.8 & 78.0 & 89.39$^{\dag}$ \\
MTB & BERT$_{large}$ & No & \textit{w/} & 71.5 & - & - \\
KnowBERT & BERT$_{base}$ & No & \textit{w/} & 71.5 & 79.3 & - \\
K-Adapter & RoBERTa$_{large}$ & No & \textit{w/} & 72.0 & - & - \\
LUKE & RoBERTa$_{large}$ & No & \textit{w/} & 72.7 & 80.6 & - \\
CR & BERT$_{large}$ & No & \textit{w/o} & 73.0 & - & - \\
RECENT & SpanBERT$_{large}$ & No & \textit{w/o} & 75.2 & - & - \\
NLI & DeBERTa$_{large}$ & No & \textit{w/} & 73.9 & - & - \\
PTR & RoBERTa$_{large}$ & Yes & \textit{w/o} & 72.4 & 81.4 & - \\
KnowPrompt & RoBERTa$_{large}$ & Yes & \textit{w/o} & 72.4 & 82.4 & - \\
FPC & RoBERTa$_{large}$ & Yes & \textit{w/o} & 76.2 & \textbf{84.9} & 89.30$^{\dag}$ \\
GenPT & RoBERTa$_{large}$ & Yes & \textit{w/o} & 74.7 & 83.4 & 89.09$^{\dag}$ \\
\midrule
\multicolumn{7}{c}{Generative Methods} \\
\midrule
SFT & T5$_{large}$ & No & \textit{w/o} & 65.16$^{\dag}$ & 75.59$^{\dag}$ & 84.75$^{\dag}$ \\
TANL & T5$_{base}$ & No & \textit{w/o} & 71.9 & - & 90.8 \\
SURE & PEGASUS$_{large}$ & Yes & \textit{w/} & 75.1 & 83.3 & - \\
GenPT & BART$_{large}$ & Yes & \textit{w/o} & 74.6 & 82.9 & 87.34$^{\dag}$ \\
GenPT & T5$_{large}$ & Yes & \textit{w/o} & 75.3 & 84.0 & 87.27$^{\dag}$ \\
\midrule
\midrule
CPTuning\textit{-s1} & T5$_{large}$ & Yes & \textit{w/o} & 76.7 & 84.5 & 91.2 \\
CPTuning\textit{-s2} & T5$_{large}$ & Yes & \textit{w/o} & \textbf{77.1} & \textbf{84.9} & \textbf{91.4} \\
\bottomrule
\end{tabular}
\vspace{-10pt}
\end{table}

\subsection{Results}
\label{sec:4.4}
Table~\ref{tab:2} presents the comprehensive evaluation results of our proposed CPTuning and the baseline methods. We begin by analyzing the results on single-relation TACRED and TACREV datasets, followed by the results on multi-relation NYT.

\textbf{Single-ralation RE.}
Among the mentioned two types of baselines, FPC demonstrated the best performance in the classification methods group, while GenPT exhibited the best results in the generative group. We note that GenPT approaches RE as a Seq2Seq text-infilling problem, similar to our approach. To some extent, this suggests the compatibility between the two tasks, which is worth exploring in future studies. 
Besides, generative methods commonly outperform the classificational, which echoes the findings in previous studies.
On the other hand, we observe that CPTuning significantly enhances the previously best performance of single-relation RE. Specifically, compared with GenPT, T5$_{large}$ fine-tuned with CPTuning\textit{-s2} achieved a 1.8 and 0.9 improvement in micro F1 scores on TACRED and TACREV, respectively. Both versions of CPTuning outperformed traditional supervised learning with MLE by a large margin. CPTuning\textit{-s2} slightly outperforms CPTuning\textit{-s1}. We attribute it to the higher informativeness of the template \textit{-s2}. \\
\indent \textbf{Multi-ralation RE.}
Since almost all baselines give up considering the EPO circumstance, we augment them with a ranking strategy to enable the comparison in multi-relation settings. Suppose an instance in NYT has $n$ annotated gold relations. As for classificational methods, we get the relation word distributions from a model’s output and select the top-$n$ verbalized relations with the highest probability mass as the prediction results. For generative methods, we require a model to generate $K$ verbalized candidate relations using beam search and then select the top-$n$ with the highest estimated likelihood as the results. 
We have two findings from Table~\ref{tab:2}. Firstly, apart from TANL and our CPTuning, the other methods show an undesirable F1 score on NYT primarily due to EPO. Additionally, CPTuning surpasses TANL on NYT by 0.4 and 0.6 points of micro F1, and the template style demonstrates little impact on the model performance.


\begin{table}[!t]
\centering
\tabcolsep=10pt
\scriptsize
\caption{
Ablation study results of CPTuning\textit{-s1}.
}
\label{tab:3}
\begin{tabular}{c|c|c|c|c|c}
\toprule
\textit{LBLS} & \textit{CTL} & TACRED & TACREV & Re-TACRED & NYT \\
\midrule
\XSolidBold & \XSolidBold & 69.9 & 81.3 & 86.4 & 85.1 \\
\CheckmarkBold & \XSolidBold & 73.8 & 81.5 & 88.8 & 87.7 \\
\XSolidBold & \CheckmarkBold & 73.0 & 81.1 & 89.3 & 87.9 \\
\CheckmarkBold & \CheckmarkBold & 76.7 & 84.5 & 91.2 & 91.2 \\
\bottomrule
\end{tabular}
\end{table}

\begin{table}[!t]
\centering
\tabcolsep=7pt
\scriptsize
\caption{
Comparison results on TACRED measured by $\mathrm{H}$-index.
}
\label{tab:5}
\begin{tabular}{c|c|c|c|c|c}
\toprule
Methods & Base Model & $\mathrm{H}@$1 & $\mathrm{H}@$5 & $\mathrm{H}@$10 & $\mathrm{H}@$20 \\
\midrule
SFT & RoBERTa$_{large}$ & 50.9 & 63.8 & 74.8 & 79.9 \\
SFT & T5$_{large}$ & 52.6 & 71.7 & 76.4 & 84.0 \\
CPTuning\textit{-s1} & T5$_{large}$ & 53.5 & 73.9 & 79.8 & 85.1 \\
\bottomrule
\end{tabular}
\end{table}

\section{Analysis}
\textbf{Ablation Study.}
We conducted an ablation study to examine the effectiveness of our designed layer-based label smoothing (LBLS) and contrastive-based learning (CTL). We remove one or both of them from the learning procedure and show the corresponding CPTuning\textit{-s1} performance in Table~\ref{tab:3}. 
Note that without both strategies, CPTuning degenerates into a supervised fine-tuning method, which performs our transformed RE and trains a model to minimize the CE loss in Eq.~\ref{eq:4}. We name this variant CPTuning$_{ce}$, and it shows notorious superiorities over the baseline SFT, especially across the three single-relation RE datasets, indicating the effectiveness of our task transformation strategy.
On the other hand, either augmenting the CE loss with LBLS - CPTuning$_{lbls}$ or attaching it with a contrastive loss CPTuning$_{ctl}$ can significantly improve the CPTuning$_{ce}$ performance on NYT, and the maximum gain is achieved by using them together, i.e., the standard CPTuning. 
We also find that using both strategies together can further enhance single-relation RE performance. All these results reveal that the two proposed non-deterministic likelihood assumptions mutually enhance each other during learning.



\begin{figure}[!t]
\centering
\includegraphics[width=1.0\linewidth]{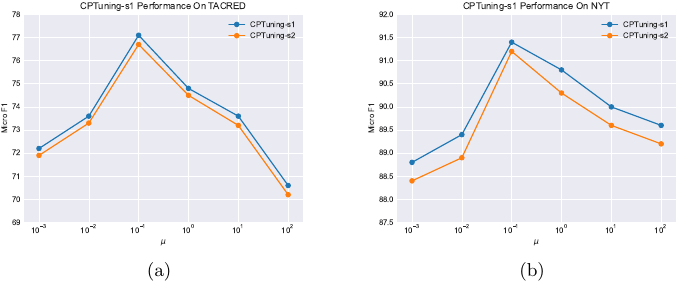}
\caption{CPTuning\textit{-s1} performance with varying $\mu$ on TACRED (a) and NYT (b).}
\label{fig:5}
\end{figure}

\textbf{Semantics Analysis of Model Outputs.}
It is worth noting that CPTuning reforms RE as a Seq2Seq text generation task, generating relation words in the target text rather than predicting numerical labels. Intuitively, it understands the semantics entailed in a verbalized relation. 
We introduce a $\mathrm{H}$-index to quantify the strength of this ability. 
Given an instance-relation pair $(s,r)$, we can select top-$M$ candidate relations $\mathcal{C}_1$ from the model's outputs according to estimated likelihood. 
Also, we can find $M$ relations $\mathcal{C}_2$ that are semantically closest to $r$ from the relation set according to semantic similarity determined by a bidirectional language model, like BERT~\cite{BERT}. We define $\mathrm{H}@M$ as the Intersection of Union (IoU) of the sets $\mathcal{C}_1$ and $\mathcal{C}_2$, which measures the consistency of likelihood estimation and semantics estimation.
Results in Table~\ref{tab:5} reveal that CPTuning is advanced in capturing the semantics feature of relation words. It assigns a higher probability mass to the candidate relation semantically closer to the gold one, which is the key for accurate multiple relations extraction.

\textbf{Effect of the Balance Factor.}
To test the influence of the balance factor $\mu$ on learning efficacy, we scale it from 0.001 to 100 and show the corresponding evaluation results of our CPTuning methods in Figure 5. We find $\mu=0.1$ is the best on both TACRED and NYT datasets. Since TACREV and Re-TACRED are essentially adapted from TACRED, we omit testing on these two datasets and set $\mu$ to 0.1 across all our experiments.



\begin{table}[!t]
\centering
\tabcolsep=8pt
\scriptsize
\caption{
Comparison results under low-resource settings.
Results marked with ${\dag}$ are from our implementation; otherwise, from \cite{GenPT}.
}
\label{tab:4}
\begin{tabular}{c|c|c|c|c|c}
\toprule
Model & \#Samples & TACRED & TACREV & Re-TACRED & NYT \\
\midrule
\multirow{3}{*}{SFT (T5)} & $N$=8 & 12.2 & 13.5 & 28.5 & 27.8 \\
& $N$=16 & 21.5 & 22.3 & 49.5 & 49.9 \\
& $N$=32 & 28.0 & 28.2 & 56.0 & 54.7 \\
\midrule
\multirow{3}{*}{GDPNet} & $N$=8 & 11.8 & 12.3 & 29.0 & 30.4\\
& $N$=16 & 22.5 & 23.8 & 50.0 & 53.1 \\
& $N$=32 & 28.8 & 29.1 & 56.5 & 53.8 \\
\midrule
\multirow{3}{*}{PTR} & $N$=8 & 28.1 & 28.7 & 51.5 & 51.6 \\
& $N$=16 & 30.7 & 31.4 & 56.2 & 54.8 \\
& $N$=32 & 32.1 & 32.4 & 61.2 & 62.4 \\
\midrule
\multirow{3}{*}{KnowPrompt} & $N$=8 & 32.0 & 32.1 & 55.3 & 58.8\\
& $N$=16 & 35.4 & 33.1 & 63.3 & 63.9 \\
& $N$=32 & 36.5 & 34.7 & 65.0 & 67.5 \\
\midrule
\multirow{3}{*}{FPC} & $N$=8 & 33.6 & 33.1 & 57.9 & 56.2 \\
& $N$=16 & 34.7 & 34.3 & 60.4 & 61.0 \\
& $N$=32 & 35.8 & 35.5 & 65.3 & 66.4 \\
\midrule
\multirow{3}{*}{CPTuning\textit{-s1}} & $N$=8 & 34.5 & 34.5 & 57.1 & 58.4 \\
& $N$=16 & 35.7 & 35.9 & 61.8 & 62.4 \\
& $N$=32 & 36.9 & 36.7 & 64.9 & 64.7 \\
\bottomrule
\end{tabular}
\end{table}

\textbf{Performance in Low-resource Settings.}
We study the CPTuning performance in low-resource settings and present the comparison results in Table~\ref{tab:4}. Following the settings of \cite{FPC}, we randomly sample $N$ instances per relation class from the training set for model learning and evaluate the trained model on the whole test set. The value of $N$ is set to 8, 16, and 32 to facilitate comparisons in our experiment. 
Firstly, although FPC and KnowPrompt achieved competitive results across varying numbers of instances, they showed somewhat unreasonable learning patterns. For example, the two methods yielded decreased F1 scores by training with fewer mislabeled instances (consider TACREV is the reannotated version of TACRED and has fewer mislabeled instances). Our CPTuning, however, achieved increasing F1 scores with the quality of a dataset improving. 
Secondly, CPTuning is less sensitive to the data scale. Taking the results on Re-TACRED as an example, increasing the instances number, CPTuning shows less gain of F1 scores than other methods.
Finally, CPTuning exhibits significant superiorities over the baseline SFT on multi-relation RE, regardless of the instance numbers.

\section{Conclusion}
In this paper, we proposed a novel contrastive-based prompt tuning method for RE, named CPTuning. 
Specifically, CPTuning reforms RE into a Seq2Seq text-infilling task using handcrafted templates. It further learns a generative language model to associate a candidate relation between two in-context entities with a probability mass above or below a threshold, corresponding to whether the relation is established.
We conducted extensive experiments on four well-known RE datasets to validate our method. The results demonstrated that CPTuning effectively overcame EPO challenges ignored in previous works, and the T5-large model fine-tuned with CPTuning exhibited state-of-the-art performance across all single and multiple relations extraction tasks.

%
%

\bibliographystyle{splncs04}
\bibliography{reference}

\end{document}